\documentclass[a4paper, 10pt, conference]{ieeeconf}      

\IEEEoverridecommandlockouts                              

\usepackage{graphics} 
\usepackage{graphicx}
\usepackage{subfigure}
\usepackage{epsfig} 
\usepackage{amsmath} 
\usepackage{cleveref}
\usepackage{color}

\title{\LARGE \bf
Real-time 3-D Mapping with Estimating Acoustic Materials
}

\author{Taeyoung Kim, Youngsun Kwon and Sung-eui Yoon
\thanks{Taeyoung Kim, Youngsun Kwon and Sung-eui Yoon are with the School of Computing, KAIST, Daejeon, South Korea. Email:
        {\tt\small taeyoungkim@kaist.ac.kr, youngsun.kwon@kaist.ac.kr, sungeui@gmail.com}}
}

\begin{document}

\maketitle
\thispagestyle{empty}
\pagestyle{empty}

\begin{abstract}
	
	This paper proposes a real-time system integrating an acoustic material estimation from visual appearance and an on-the-fly mapping in the 3-dimension. The proposed method estimates the acoustic materials of surroundings in indoor scenes and incorporates them to a 3-D occupancy map, as a robot moves around the environment. To estimate the acoustic material from the visual cue, we apply the state-of-the-art semantic segmentation CNN network based on the assumption  that the visual appearance and the acoustic materials have a strong association. Furthermore, we introduce an update policy to handle the material estimations during the online mapping process. As a result, our environment map with acoustic material can be used for sound-related robotics applications, such as sound source localization taking into account various acoustic propagation (e.g., reflection).
\end{abstract}

\section{INTRODUCTION}

Thanks to wide adoption of AI speakers (e.g., Amazon Alexa), attentions and
demands for mobile robots that use sound information to interact with people
are increasing recently.   The crucial function required for this kind of
interactions between human and robots is to distinguish the specific sound
source and to know where the sound is emitted.
These tasks  are known as  sound source separation (SSS) and sound source
localization (SSL) \cite{brandstein2013microphone}, respectively.  

The understanding of physical properties of sound is necessary for the robots
to properly process the sound information.  In the case of indoor environments,
the sound propagation reaches the robot by interacting with various objects and
structures of the indoor environment.  In order to perform the SSS or SSL, it
is important to reconstruct 3-D geometric information of indoor rooms and grasp
their acoustic materials.

Various sensors such as LIDAR \cite{moosmann2011velodyne} or RGB-D camera
provide a high amount of sensor data to build the 3-D representation of 
indoor scenes.
Point clouds captured by RGB-D sensor consist of the geometric points with
color observations, and are tend to take a high amount of the data, hindering
efficient processing in practice.
For this reason, various occupancy mapping structures based on a
grid-~\cite{moravec1996robot} and an octree-based
approaches~\cite{hornung2013octomap} were proposed. These methods build the
3-D representation of an environment consisting of the volumes with occupancy
states and colors of the surrounding objects.

Sound-related studies in the robotics area can be improved by considering
acoustic material that affect sound propagation.  The acoustic material
describes the interaction between surface and incident sound wave, and it is
generally used for simulating immersive sound generation in the AR/VR
application.  However, estimating the accurate acoustic material is an
expensive task. Recent SSL papers considering reflection
\cite{an2018reflection} and diffraction \cite{an2018diffraction} only use the
3-D geometry of indoor rooms for the real-time processing, without taking into
account acoustic material information.

\begin{figure}
	\centering
	\subfigure[An example scene]{\includegraphics[width=0.25\textwidth]{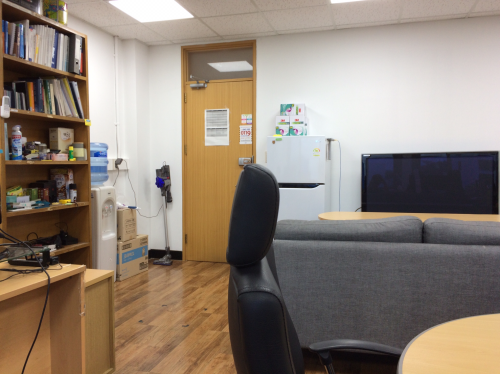}}
	\subfigure[Grid-map (color)]{\includegraphics[width=0.23\textwidth]{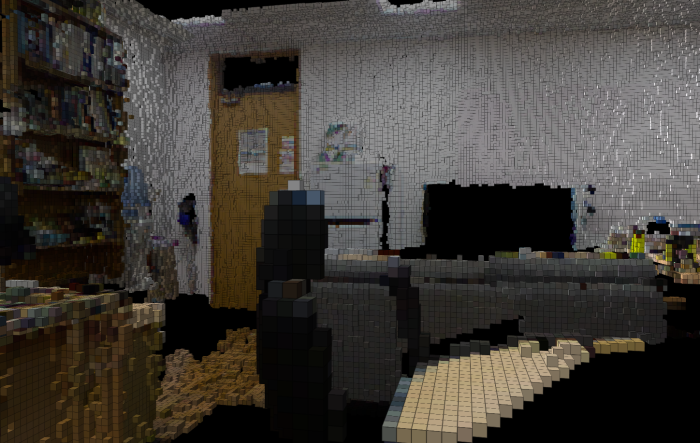}}
	\subfigure[Grid-map (acoustic material)]{\includegraphics[width=0.23\textwidth]{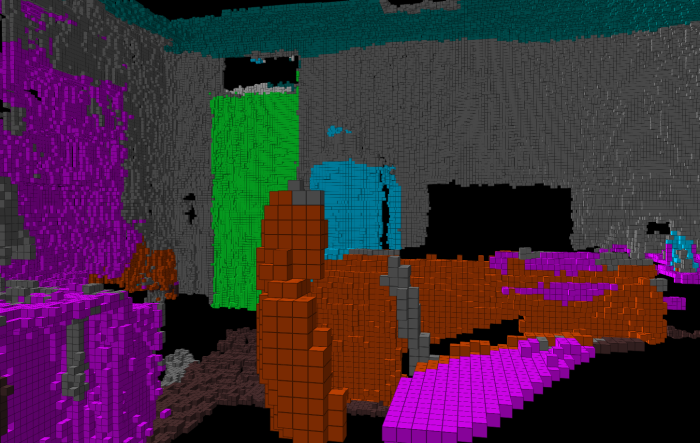}}
	\caption{An example of result of our integrated, real-time system between the
		3-D mapping and acoustic material estimation. 
		Different colors of (c) represent different acoustic materials (Fig.~\ref{fig:acoustic_material_histogram}).	}
	\label{fig:result_overview}
\end{figure}

Because accurate estimation of acoustic material is difficult, several
approaches \cite{schissler2017acoustic, kim2019immersive} were proposed for the
approximate estimation using visual appearance without sound information.
These methods use RGB or RGB-D images to perform the acoustic material
estimation and the 3-D reconstructions.  Due to the 3-D mesh reconstruction
from such images, it typically takes several hours to build a whole environment
map.  In this paper, we propose a new system that integrates real-time 3-D
mapping and the acoustic material estimation, which enables practical benefits
to sound-related applications with real-time performance.

\begin{figure*}
	\centering
	\includegraphics[width=\textwidth]{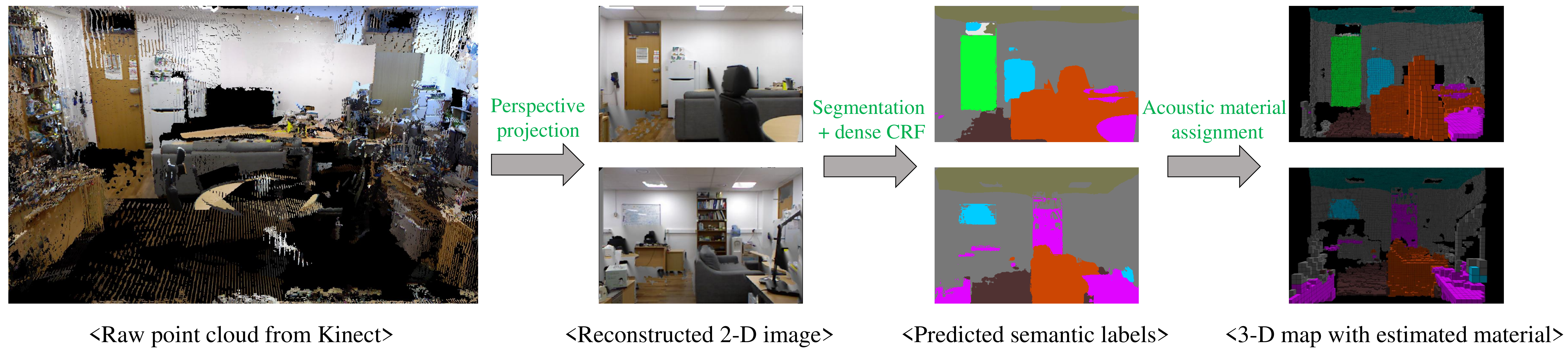}
	\caption{Pipeline of our framework.
	}
	\label{fig:overview}
\end{figure*}

\section{RELATED WORK}

\subsection{Room acoustic modeling}
Room acoustic modeling describes the behavior of sound propagation in a closure
room. This modeling is commonly used to simulate sound propagation in a room,
including properties of sound such as reflection, scattering, and diffraction.
Recent studies classify these models in two ways: wave-based and geometrical
acoustics (GA) \cite{vorlander2008auralization}. Wave-based methods directly
solve a wave equation numerically to provide accurate sound simulation.
However, this approach requires an expensive computational cost
\cite{raghuvanshi2009efficient}. In GA methods, sound signal is assumed to
propagate as rays that ignore the wave properties \cite{kuttruff2016room}. This
assumption is valid at high-frequencies when the wavelength of sound is short
compared to the interacted surface. Due to the ray assumptions, GA methods get
a less accurate result, but show a faster processing speed than the wave-based approaches.

In practice, GA methods are widely used to simulate sound propagation in an
indoor scene. To perform GA simulation, it is important to obtain an accurate
geometric model of the room and the acoustic material of that model. The
acoustic material describes how incident sound interacts with the surface
\cite{schissler2017acoustic}. It generally represents how the sound is
reflected, scattered, and transmitted through the surface of the geometric
model. The commonly used representation of acoustic material is
a frequency-dependent absorption
coefficient that describes the fraction of incident sound absorbed with each
reflection \cite{kuttruff2016room}. This property is commonly measured in
reverberation chambers according to ISO 354 \cite{ISO354}. However, estimating
the acoustic characteristic of every surface in the room using ISO 354 is
impractical. For this reason, we propose a new framework estimating the
acoustic material from the raw point cloud data obtained by a mobile robot.

\subsection{Real-time occupancy mapping}

Automatic systems using robots require understanding their workspace to perform various tasks. In the robotics field,
various occupancy maps have been studied to represent the geometry information and provide the 3-D representation of the surroundings.
As one of the mapping algorithms, grid-based representations~\cite{moravec1996robot, hornung2013octomap} are widely used 
in many systems such as RGB-D SLAM~\cite{endres20133} and sound source localization~\cite{an2018reflection}.
Compared to the learning-based approaches~\cite{doherty2017bayesian, guizilini2016large},
the grid-based methods provide real-time updates to handle the huge amount of
data efficiently that a sensor captures on-the-fly, as supported by recent
work~\cite{kwon2019super}.  Based on the benefit in practical usage, we opt to
use the grid-based approach to manage the estimated acoustic material and
represent the environment with them in real-time.

\section{METHODOLOGY}

\subsection{Overview}
Our proposed sturcuture for integrating a grid-based 3-D map representation
with acoustic material estimation is depicted in Fig.
\ref{fig:overview}.

Using a mobile robot equipped with a RGB-D sensor, we capture the point cloud
data sampled from the given environment.  Typical mapping methods perform the
update of occupancy information for the geometric representation of an
environment.  In addition to the occupancy data, our method handles the
acoustic material information acquired from the image-based estimation stage
before the occupancy updates.
In this paper, the material estimation is performed by utilizing the ResNet
architecture \cite{he2016deep} to recognize semantic labels of the current
frame.  In order to assign the acoustic property, objects with similar semantic
labels are considered to have an identical characteristic.
After the estimation process, we insert both the occupancy and the estimated
acoustic material information to the grid-based map.

\subsection{Acoustic material estimation}

There are a few studies to estimate the acoustic material using the visual cue.
Schissler et al. \cite{schissler2017acoustic} assume that there is a strong
relationship between visual appearance and its acoustic property. This method
estimates the visual appearance of the 3-D mesh using material classification
CNN \cite{bell2015material} and matches the estimated visual material to the
acoustic material of various type of architectural components. Kim et al.
\cite{kim2019immersive} also adopted a CNN-based acoustic material estimation
from the RGB-D image.
These methods require additional costs to reconstruct 3-D models from the
image, which is used to estimate the acoustic property. 
This cost could be very high, i.e., 5 minutes~\cite{kim2019immersive}, which is unsuitable for real-time robotic applications.

In this paper, we propose a different method for estimating the acoustic
material from the 3-D point cloud obtained by a robot sensor. Our approach
utilizes a well-known grid-based 3-D map and 
performs a simple perspective projection to assign the acoustic property of the
environment directly to the map without having expensive reconstruction steps.

To perform the material estimation from the point cloud, we first reconstruct
the image from the sensor data. 
Each point with RGB color obtained from the Kinect sensor locates in the sensor coordinates.
In the $XYZ$ coordinate system where $Z$ represents depth from the sensor, we
reconstruct an image by performing a perspective projection of each measured
point on the image plane located in a distance $F$ away from the current sensor
position.
Suppose that the camera coordinate is $(X_c, Y_c, Z_c)$.  Each point $p_i$
located to $(X_i, Y_i, Z_i)$ is then projected to $(X_i', Y_i')$ of the
two-dimensional image
through the \Cref{eq:perspective_projection}.

\begin{equation}
\label{eq:perspective_projection}
\begin{matrix}
X_i' = (X_i - X_c) \times (F/Z_i) + X_c,\\ \\
Y_i' = (Y_i - Y_c) \times (F/Z_i) + Y_c.
\end{matrix}
\end{equation}

We now estimate the acoustic material from the reconstructed image using the
widely used CNN to predict visual appearance. As mentioned in
\cite{schissler2017acoustic, kim2019immersive}, estimating the visual aspect of
an image can be considered as a kind of semantic segmentation.  In general,
semantic segmentation tasks classify and separate various objects in the scene.
However, there are fewer types of objects that affect sound propagation in indoor scenes.
Some of them include furniture and
ceiling, so most labels (e.g., waterfall and airplane) that are not related
to the indoor setting or small objects need to be ignored out.  

We utilize the pre-trained network with
MIT ADE20k dataset \cite{zhou2018semantic, zhou2017scene} that is the largest
for semantic segmentation and scene parsing. This network is trained for 150
objects, but we select only eight labels (Table~\ref{table:matching}) for the final result because of the
simplicity of the acoustic property in the indoor environment.
In order to assign the accurate acoustic material to each point cloud, we
apply dense CRF \cite{krahenbuhl2013parameter} to predict a label at every
pixel. 
CRF consists of two energy function, the unary term, and the pairwise term. In general, the probability of estimated label is used to the unary term. The pairwise term considers the similarity of the location and color value of pixels.
We use the details of the energy function 
of CRF from \cite{bell2015material}, which uses the LAB color space for the pairwise energy term.

\begin{table}[t]
	\caption{Detected acoustic materials matching to different object types.}
	\begin{center}
		\begin{tabular}{|c|c||c|c|}
			\hline
			\textbf{Object} & \textbf{Material} & \textbf{Object} & \textbf{Material} \\ \hline
			Wall & Concrete & Floor & Linoleum \\ \hline
			Ceiling  & Plywood & Window & Thick glass \\ \hline
			Furniture & Wood & Door & Wood panel \\ \hline
			Electronics  & Plastic &  Chair& Carpet \\ \hline
		\end{tabular}
	\end{center}
	\label{table:matching}
\end{table}

As the next step, we match each semantically classified object to one of
available measured materials, whose acoustic materials are measured in a
controlled setting~\cite{ermann2015architectural, vorlander2007auralization}.
\Cref{table:matching} shows the corresponding acoustic materials that match the
eight selected labels.  After assigning the material to all pixels in the
reconstructed 2-D image, we pass this label information to the original raw
point cloud. To realize this process,
we maintain the relationship between the 3-D point cloud and the location of
corresponding pixels, when we project the point cloud into the 2-D image using
\Cref{eq:perspective_projection}. 
The point cloud with the information
of acoustic material is passed to the next step for mapping the current
environment.

\subsection{Real-time 3-D mapping with acoustic material}

An occupancy grid discretizes the world into a set of the cells containing
the occupancy information such as the occupied or free states.  
Each cell of the map has a
volumetric representation, voxel, of the partitioned space where 
multiple points of a point cloud can be assigned to. Unlike the
traditional occupancy grid, our system handles a point cloud consisting of
the points with the color measurements as well as the estimations of
acoustic material.  Therefore, we modify a
design of the cell with the occupied state to hold this data about the
material.

In this work, we use a mobile robot to construct our occupancy grid with
acoustic material for the applications related to sound.  When the robot
roams around the workspace, our approach incrementally updates the various
information, i.e., occupancy, color, and estimated sound material, of each
cell associated with the measured points on-the-fly. We adopt the policy for
updating the occupancy probability and the color based on the outcome of the
prior work~\cite{hornung2013octomap}. However, in the case of the acoustic
material, we cannot easily integrate the estimated labels to a single value
on a simple manner like averaging the estimated labels.

\begin{figure}
	\centering
	\includegraphics[width=\columnwidth]{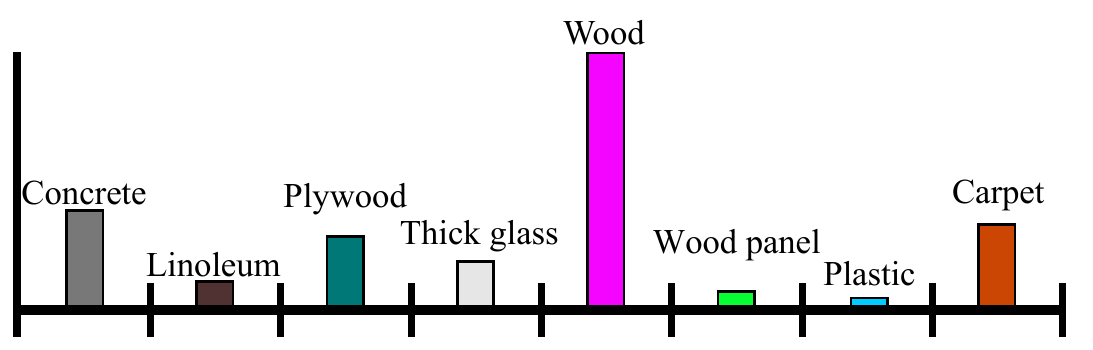}
	\caption{An example of our acoustic material histogram. Colors assigned to different materials will be used for material visualization later (e.g., Fig.~\ref{fig:whole_res}).}
	\label{fig:acoustic_material_histogram}
\end{figure}

We, therefore, propose a histogram-based method to efficiently update and
determine the acoustic material of the cells for on-the-fly mapping.  In our
occupancy grid, we design each cell to have its own histogram, which counts
the estimated
labels of the material to each bin
(Fig.~\ref{fig:acoustic_material_histogram}).  Our method then determines
the material of a cell as the most frequently-appearing  estimation to the
acoustic material at the points mapped to the cell.
For the example shown in Fig.~\ref{fig:acoustic_material_histogram}, our
method selects the wood as the acoustic material of the cell, since the
estimated label has the largest density among the identifications of the
acoustic material.  This histogram-based approach is useful to manage and
filter the noisy estimations from the points assigned to the cell.

\section{IMPLEMENTATION}

\begin{figure}
	\centering
	\subfigure[Our robot]{\includegraphics[width=0.09\textwidth]{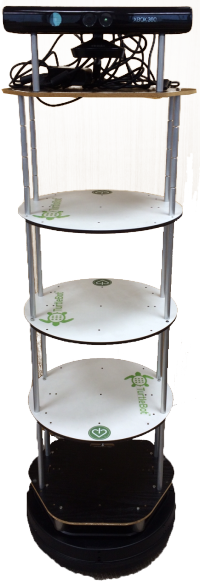}}
	\subfigure[Tested indoor scene]{\includegraphics[width=0.385\textwidth]{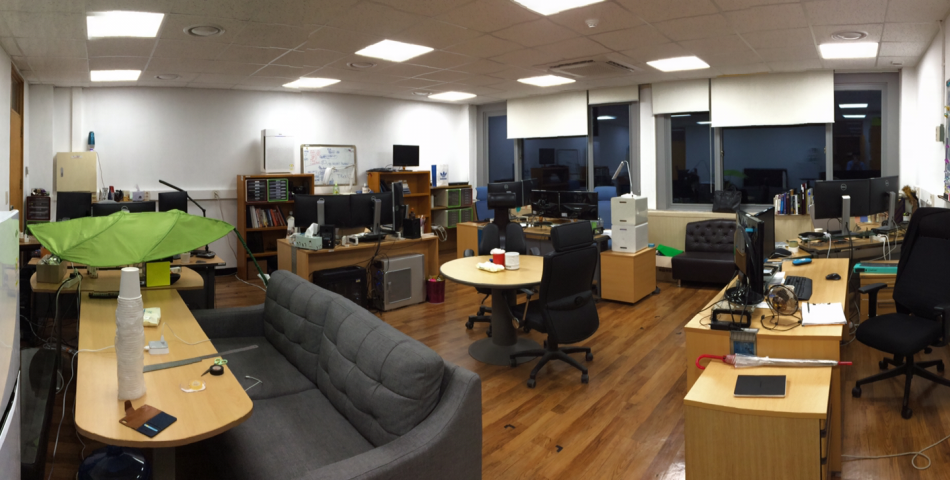}}
	\caption{Our tested robot and scene}
	\label{fig:env}
\end{figure}

Fig. \ref{fig:env} shows our tested robot used for estimating the room structure. This robot is based on a Turtlebot2 equipped with Microsoft Kinect sensor that located 1.08m from the ground. The sensor captures 30k points each frame.
We tested our proposed system in the common office rooms, whose
dimensions are 6.7$\times$6.8$\times$2.5m. We acquire the
3-D point cloud of the room as the robot moves along a given path.

\begin{figure}
	\centering
	\subfigure[Reconstructed 2-D image]{\includegraphics[width=0.25\textwidth]{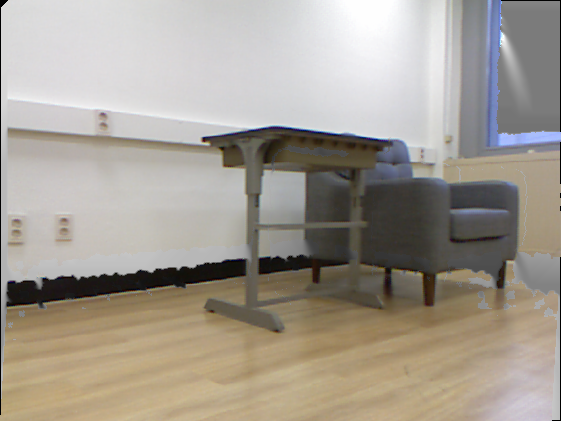}}
	\subfigure[Initial semantic segmentation]{\includegraphics[width=0.23\textwidth]{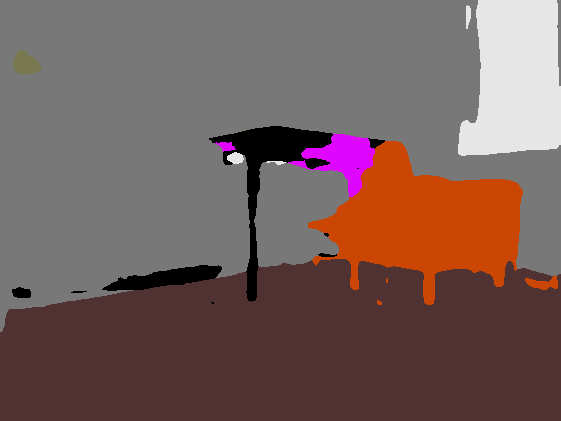}}
	\subfigure[After dense CRF]{\includegraphics[width=0.23\textwidth]{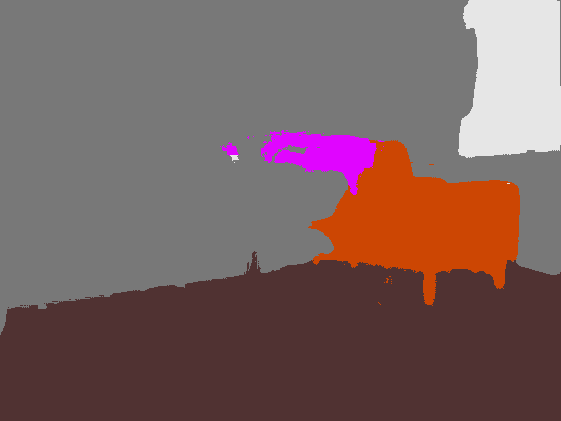}}
	\caption{Segmentation result of reconstructed image}
	\label{fig:seg_result}
\end{figure}

We reconstruct the 2-D RGB image from the captured point cloud.  Fig.
\ref{fig:seg_result}-(a) shows the reconstructed 2-D image from the raw point
cloud.  Due to the properties of the point cloud from the sensor, the
reconstructed 2-D image includes empty pixels, i.e., holes,  that are
unobserved from the sensor.  Since this unobserved points degrade the
performance of image-based semantic segmentation, we apply a mean filter to
fill these area with the average color of adjacent pixels.

The reconstructed images are used as the input of semantic segmentation
network. We use the pretrained ResNet \cite{he2016deep} architecture with MIT
ADE20k dataset \cite{zhou2018semantic, zhou2017scene} for semantic
segmentation.  Since only eight classes are used to estimate the acoustic
material of indoor objects among the 150 classes, the unclassified pixels are
represented by unknown labels. Fig. \ref{fig:seg_result}-(b) shows the
initial segmentation result that contains the unclassified pixels represented
by the black color. In order to fill this area, we apply the dense CRF
\cite{krahenbuhl2013parameter} to assign a label to every pixel as shown in
Fig. \ref{fig:seg_result}-(c).  
We then add this information to the original point cloud and
pass it to the environment mapping stage.

Our occupancy map updates the representation of the geometry as well as the
acoustic material from the point cloud.  We implemented the occupancy grid
using the library published in the prior work~\cite{kwon2019super}.

\section{RESULT AND DISCUSSION}

\begin{figure*}
	\centering
	\small
	\begin{tabular}{cccc}
		\includegraphics[height=0.1\textheight]{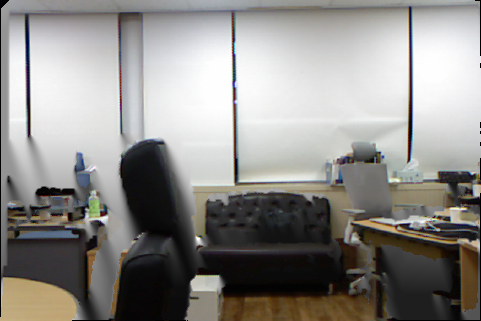}&
		\includegraphics[height=0.1\textheight]{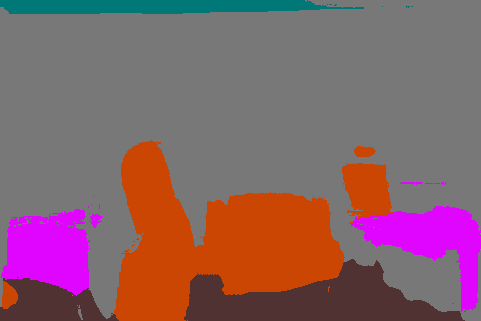}&
		\includegraphics[height=0.1\textheight]{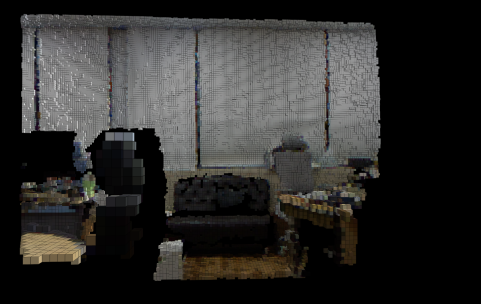}&
		\includegraphics[height=0.1\textheight]{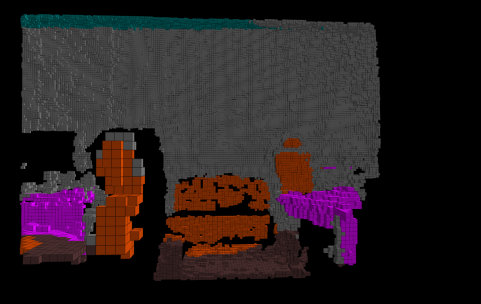}\\
		
		\includegraphics[height=0.1\textheight]{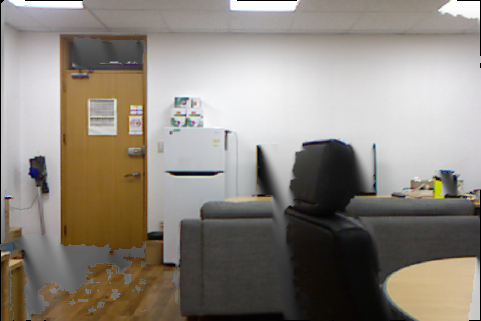}&
		\includegraphics[height=0.1\textheight]{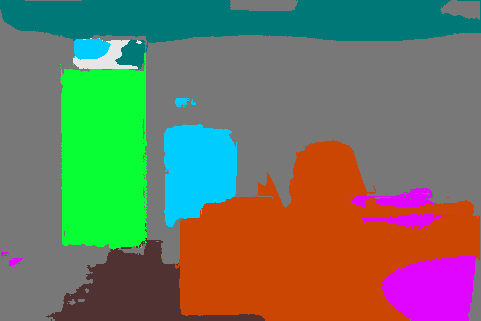}&
		\includegraphics[height=0.1\textheight]{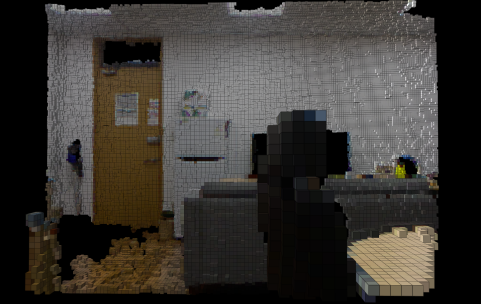}&
		\includegraphics[height=0.1\textheight]{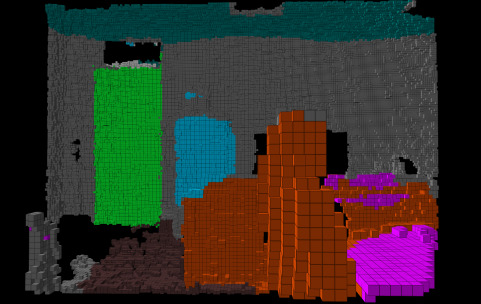}\\
		
		\includegraphics[height=0.1\textheight]{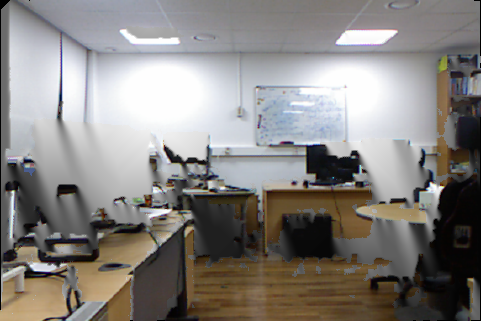}&
		\includegraphics[height=0.1\textheight]{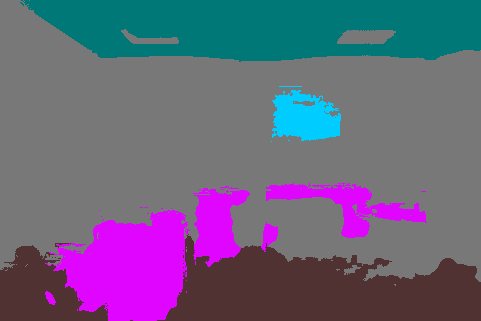}&
		\includegraphics[height=0.1\textheight]{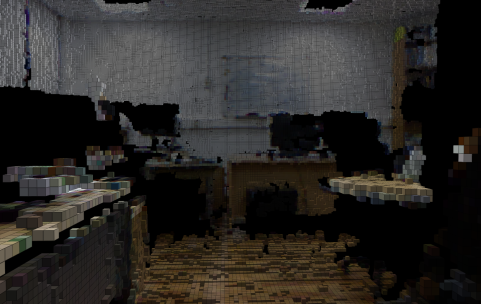}&
		\includegraphics[height=0.1\textheight]{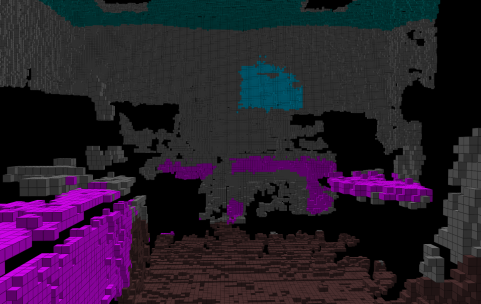}\\
		
		\includegraphics[height=0.1\textheight]{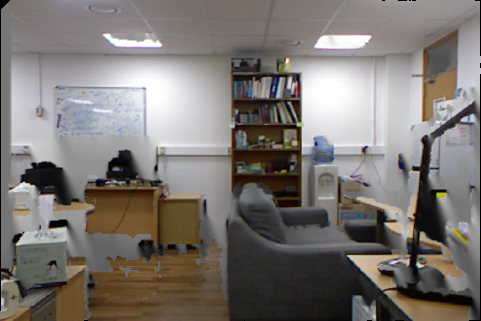}&
		\includegraphics[height=0.1\textheight]{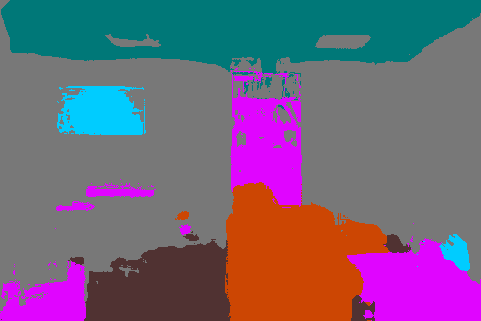}&
		\includegraphics[height=0.1\textheight]{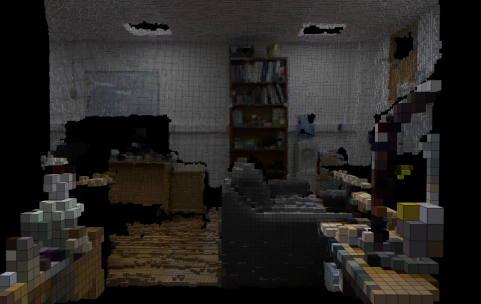}&
		\includegraphics[height=0.1\textheight]{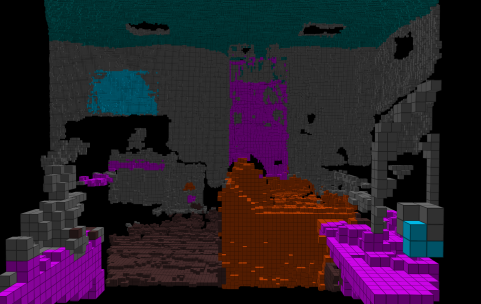}\\
		(a) Reconstructed image&(b) Segmented results&(c) 3-D color map&(d) 3-D labeled map
	\end{tabular}
	
	\caption{Example of our proposed system}
	\label{fig:whole_res} 
\end{figure*}

Fig. \ref{fig:whole_res} shows the results of our proposed system at the
tested room.  
Fig.  \ref{fig:whole_res}-(a) and -(b) show the reconstructed image using
the mean filter to remove empty pixels and the result of semantic segmentation
with dense CRF, respectively. After the segmentation, we assign the material
information to each point according to the matching strategy summarized in
Table \ref{table:matching}, as shown in the Fig. \ref{fig:whole_res}-(d). 
In our tested scene, various objects generate occlusion producing a higher frequency of unobserved
areas. Particularly, specular surfaces or transparent objects influence a
negative effect on the point cloud sensing and cannot receive data, as shown in the Fig. \ref{fig:whole_res}-(a) such as a monitor and glass.
These problems
produce in poor segmentation results, but it is possible to assign reasonable
acoustic material to dominant objects in the scene.

Our proposed method mainly depends on the performance of semantic segmentation.
Using the point cloud data rather than RGB-D images makes it easier to
reconstruct 3-D environment map, but there is a problem that it is hard for the
employed sensor to acquire points for specular or transparent
objects as mentioned above.

Commonly used networks for semantic segmentation
learn from photographs taken from the human's view. 
However, in the case of images obtained using a robot, the point of view of the camera also
locates low. 
We found that this difference of views causes the misprediction of the semantic structure. To solve this misprediction, we increased the height of the robot camera's view, as shown in Fig. \ref{fig:env}-(a).

We expect that this problem also can be solved by learning networks from the various
data acquired by robots. However, dataset construction from the robot sensor is
an expensive task. In this reason, we plan to design a new network and dataset
for material estimation as a part of future work.

\section{CONCLUSION}

This study aimed to design an on-the-fly system that predicts acoustic material
with visual appearance and conducts an environment mapping using a 3-D point
cloud, based on the assumption that visual information is highly correlated 
with
acoustic material. The material of each point is estimated by reconstructing
the 2-D image from the 3-D point cloud received from the Kinect sensor and
assigning an acoustic information that matches each semantic object by the
result of segmentation on the reconstructed image.  Finally, the overall
environment map has the estimated material as well as occupancy
information for each voxel.

Since the estimation is based on the visual information, it has approximate
acoustic information. Also, because this estimation process highly depends on
the performance of the segmentation network and the quality of the
reconstructed 2-D image, the segmented result may include mispredicted labels.
However, the productivity of sound-related researches can be increased by
automating the assignment of acoustic materials, which can be improved by
considering various other acoustic properties.

\addtolength{\textheight}{-12cm}   




{
\bibliography{ms}
\bibliographystyle{IEEEtran}
}

\end{document}